\documentclass{article} 
\usepackage{iclr2025_conference,times}

\usepackage{multirow}
\usepackage[table]{xcolor}

\iclrfinalcopy
\definecolor{highlight2}{RGB}{255,242,170} 
\definecolor{highlight}{RGB}{255,220,220} 

\usepackage{url}
\usepackage[utf8]{inputenc}
\usepackage{newunicodechar}
\newunicodechar{−}{\ensuremath{-}} 
\usepackage{makecell} 

\usepackage{multirow}
\usepackage{array}

\usepackage{longtable}

\usepackage{amsmath,amsfonts,bm}

\def\eqref#1{equation~\ref{#1}}

\def\1{\bm{1}}

\DeclareMathAlphabet{\mathsfit}{\encodingdefault}{\sfdefault}{m}{sl}
\SetMathAlphabet{\mathsfit}{bold}{\encodingdefault}{\sfdefault}{bx}{n}

\usepackage{hyperref,booktabs,enumitem,xcolor}
\usepackage{graphicx}

\title{From Canon to Nikon: Camera Domain Shift and Potential Equity Implications in Retinal AI} 

\author{Lisa Y.W. Tang\\
University of British Columbia - Faculty of Medicine}

\begin{document}

\maketitle

\begin{abstract}
Diabetic retinopathy is a leading cause of preventable blindness, yet access to regular screening remains limited worldwide. AI offers a scalable approach, but its performance can be sensitive to image quality and acquisition conditions. In a Brazilian cohort of 8,524 patients, retinal images acquired with the Nikon NF5050 were more than twice as likely to have inadequate image quality compared with those acquired with the Canon CR-2, despite refurbished Canon systems costing more than twice as much as Nikon systems. This potential cost–quality trade-off highlights an important consideration for the reliability of AI systems deployed across heterogeneous imaging systems. Using a frozen pretrained retinal foundation model, we systematically evaluate whether camera-associated performance differences persist across 12 retinal disease classification tasks, four input resolutions, and five downstream classifiers. We find that camera-associated performance differences vary substantially across tasks, and that neither higher input resolution nor classifier choice consistently mitigates cross-camera variation. These findings identify camera platform as a potentially overlooked source of performance variability, with implications for the safety and reliability of retinal AI when deployed across heterogeneous imaging systems. We release all source code, evaluation scripts, and per-task results on on \href{https://anonymous.4open.science/r/Canon2Nikon/README.md}{GitHub} to support reproducible benchmarking. 
\end{abstract}

\textbf{\large{1 Introduction}}

Diabetic retinopathy (DR) is a leading cause of preventable blindness, affecting more than 100 million people worldwide \citep{teo2021global}. The burden is particularly high in low- and middle-income countries, where access to ophthalmic care and regular screening remains limited. Automated analysis of retinal fundus photographs offers a scalable approach to screening, but models that perform well under controlled settings  degrade when deployed across different clinical environments \citep{ting2017development,beede2020human}.

One important source of distribution shift is the fundus camera. Differences in optics, sensors, acquisition protocols, and image-processing pipelines can alter the visual characteristics of retinal photographs and affect model performance. Han et al., for example, reported substantial variation in referable DR detection across fundus-camera models, with performance values ranging from 0.818 to 0.978 \citep{han2022validating}. Such variation is particularly relevant to screening programs, where models may encounter imaging systems that were not represented during development. 

Brazil provides a relevant setting for studying this problem. Its geographically distributed screening infrastructure relies on heterogeneous imaging equipment, including both high-end desktop systems and lower-cost alternatives. In a Brazilian dataset (BRSET) \citep{brset2024} involving 8,524 patients, retinal images acquired with the Nikon NF5050 were more than twice as likely to have inadequate image quality compared with those acquired with the Canon CR-2, despite Canon systems costing approximately 2 times as much when refurbished
\footnote{In the latest local refurbished market, Nikon NF5050 is listed at approximately R\$35,000 (U\S6,000--6,500), compared with approximately R\$95,000 (US\$10,000) for a Canon CR-2 system \citep{NikonNF505Brazil,OftalmoTradeCanonCR2}.}. This potential cost–quality trade-off highlights an important consideration for reliable and equitable AI deployment.

Prior studies have examined cross-camera generalization in retinal image analysis, including comparisons between desktop and portable imaging systems \citep{han2022validating,He2023Crosscamera}. While these studies have demonstrated domain shift between desktop and portable systems, less is known about variation across different desktop camera platforms and how this interacts with input resolution and downstream model choice. Moreover, existing studies have largely focused on DR, leaving unclear whether camera-related performance differences extend across other retinal diseases. 

We address this literature gap by using RETFound-Green (RFG). RFG is a recently proposed retinal foundation model that is 400x more computationally efficient than its predecessor, substantially reducing the associated CO$_2$ emissions \citep{engelmann2025training}. Building on the authors' emphasis on computational and environmental efficiency, we investigate whether RFG can also reduce the need for costly fine-tuning by using it as a frozen feature extractor. Freezing the model allows us to reuse the same learned retinal representation across our experimental settings, avoiding repeated fine-tuning across the full set of input-resolution and downstream-classifier combinations. This substantially reduces the computational cost and associated CO$_2$ emissions of the experimental evaluation while allowing us to systematically assess the effect of these design choices.


To this end, we evaluate 12 clinical tasks spanning distinct pathological phenotypes, providing a broader assessment than a single DR endpoint. To ensure a fair and leakage-free comparison, we use patient-level partitions throughout, in contrast to the original BRSET benchmark \citep{brset2024}, and report 95\% bootstrap confidence intervals.

\vspace{1em}
\textbf{\large{2 Methods}}

\paragraph{Dataset.}

The Brazilian Multilabel Ophthalmological Dataset (BRSET) consists of retinal fundus photographs collected between 2010 and 2020 from three outpatient ophthalmological centers in São Paulo, Brazil \citep{brset2024}. The dataset contains 8,524 patients, of whom approximately 16\% had diabetes. Approximately 65\% of images were acquired using the Canon platform; the remaining were acquired using Nikon NF5050. The two systems differ in their imaging hardware and acquisition and processing characteristics, providing a natural setting for evaluating camera-associated domain shift. Images were labeled by retinal specialists according to predefined protocols. Image quality was assessed using four criteria: focus, illumination, image field, and artifacts. Each criterion was classified as satisfactory or unsatisfactory, and an image was considered ``inadequate'' if any criterion was unsatisfactory.

\paragraph{Retinal foundation model.}

We intentionally adopt a frozen feature extraction paradigm to leverage 
RETFound-Green (RFG) \citep{engelmann2025training}, a recently proposed retinal foundation model, as a frozen feature extractor for all downstream experiments. RFG is a 22.2-million-parameter vision transformer that was initialized with weights from DINOv2 \citep{oquab2024dinov2} and subsequently pretrained on retinal images. It uses a ``self-supervised Token Reconstruction'' objective to  capture global image information by incorporating four special learnable vectors to interact with the image patch tokens through self-attention. During training, a frozen copy of the model processes the original image to produce target tokens, while a second copy receives a noisy version of the same image and is trained to reconstruct these tokens. The reter tokens therefore participate in the self-attention process and are learned together with the rest of the model during retinal pretraining.  RFG was pretrained on 75,000 retinal fundus photographs using less than \$100 of computational resources, 400× less compute than its predecessors \citep{engelmann2025training}.



\paragraph{Experimental setup.}

We evaluated five downstream classifiers: support vector machine (SVM), k-nearest neighbors (kNN), histogram-based gradient boosting (HGB), logistic regression (LR), and random forest (RF). These classifiers span linear, margin-based, instance-based, and tree-based approaches. All classifiers were evaluated using the same RFG embeddings and under the same patient-level data partitions. The experimental factors were camera platform, input resolution, classifier, and diagnostic task. We evaluated 12 retinal disease classification tasks across four input resolutions (224, 392, 672, and 952 pixels) and five classifiers.

\paragraph{Camera-domain evaluation.}

To evaluate camera-associated domain shift, we separated evaluation by camera platform rather than pooling images from both cameras. The Canon cohort was used as the source domain and the Nikon cohort as the target domain. Models were trained and internally selected using source-camera data and subsequently evaluated on the held-out target-camera data without further model fitting. This source-to-target evaluation measures whether a classifier trained on one camera platform maintains performance when applied to images acquired with another platform. Within-camera performance was also evaluated to distinguish general classification performance from cross-camera degradation.

\paragraph{Input resolution.}

We evaluated four input resolutions: 224, 392, 672, and 952 pixels. The 224$\times$224 resolution provides a commonly used baseline, while 392$\times$392 corresponds to the input resolution used for RFG pretraining. The 672$\times$672 resolution provides an intermediate, non-dyadic scaling relative to 392$\times$392, allowing us to assess whether performance changes smoothly with resolution or depends on the particular resampling scale. Finally, 952$\times$952 was selected to maintain compatibility with the model's 14$\times$14 patch structure. All four resolutions are divisible by 14, corresponding to patch grids of 16$\times$16, 28$\times$28, 48$\times$48, and 68$\times$68, respectively.

\paragraph{Model selection and evaluation.}

For each diagnostic task and input resolution, classifier hyperparameters were selected using only the source-camera training and validation data. The selected model was then evaluated on the held-out source-camera test set and the target-camera test set without further tuning. For binary classification tasks, performance was measured using area under the receiver operating characteristic curve (AUROC). For DR staging, quadratic weighted kappa (QWK) was used as the primary metric. Performance estimates are reported with 95\% bootstrap confidence intervals. All data partitions were performed at the patient level to ensure that images from the same patient did not occur across training, validation, and test partitions.

\vspace{1em}
\textbf{\large{3 Results}}

\textbf{Image quality differs across camera platforms.} Image quality differed between the two camera platforms (Table~\ref{tab:qi}). Inadequate images were more frequent in the Nikon NF5050 cohort than in the Canon CR-2 cohort (18.6\% vs.\ 8.8\%; $\chi^2(1)=328.19$, $p<0.001$, Cramér's $V=0.142$). The largest difference occurred for image-field adequacy, which was classified as inadequate in 15.84\% of Nikon images compared with 4.76\% of Canon images ($V=0.188$, $p<0.001$). In contrast, focus and artifact adequacy showed little association with camera platform ($V=0.018$ and $V=0$, respectively). Illumination showed a statistically significant but weak association ($V=0.028$, $p<0.001$). These results indicate that camera platform in BRSET is associated not only with image acquisition characteristics but also with image quality, an important consideration when interpreting subsequent camera-domain comparisons.

\begin{table}[b!]
    \centering
    \small
    \caption{Percentage of images classified as inadequate for each image-quality criterion by camera platform, with corresponding camera--quality association statistics. Cramér's $V$ quantifies the strength of association between camera platform and each quality criterion.}
    \label{tab:qi}
    \begin{tabular}{l|c|cccc}
        \toprule
        Camera & Overall & Focus & Illumination & Image field & Artifacts \\
        \midrule
        Canon CR
        & 8.79\%
        & 3.57\%
        & 0.67\%
        & 4.76\%
        & 0.35\% \\
        
        Nikon NF5050
        & 18.60\%
        & 2.90\%
        & 0.23\%
        & 15.84\%
        & 0.35\% \\
        \midrule
        Cramér's $V$ ($p$)
        & 0.142  ($<0.001$)
        & 0.018 (0.065)
        & 0.028 ($<0.001$)
        & 0.188 ($<0.001$)
        & 0.000 (1.000) \\        
        \bottomrule
    \end{tabular}
    
\end{table}

\textbf{Camera-associated performance varies across disease tasks.} Camera-associated differences in performance were heterogeneous across diagnostic tasks. Table~\ref{tab:summary} summarizes our experimental results, while all 240 comparisons (12 tasks × 4 resolutions × 5 classifiers) and corresponding 95\% confidence intervals are provided in the Appendix. Several tasks showed little difference between the Canon and Nikon cohorts. For example, AUROC was similar for diabetic retinopathy (0.940 vs.\ 0.946), scar (0.964 vs.\ 0.955), and macular edema (0.952 vs.\ 0.952). In contrast, larger reductions were observed for hemorrhage (0.804 vs.\ 0.569), hypertensive retinopathy (0.864 vs.\ 0.690), and DR staging (0.692 vs.\ 0.527). Other tasks showed intermediate differences, including AMD (0.976 vs.\ 0.909) and nevus (0.840 vs.\ 0.795). Thus, camera-associated performance differences were not uniform across disease phenotypes. The same camera shift that produced substantial degradation for some tasks had little apparent effect on others.

\begin{table}[t]
\centering  
\caption{Best-performing classifier (selected by internal validation AUROC, or QWK for DR staging) per task, with internal (train and test images acquired using Canon) and external (Canon for training and Nikon for test) performance at the selected resolution. Values show point estimate with 95\% bootstrap confidence interval. Cell color encodes the magnitude of the performance drop relative to the Canon reference split: yellow for drops up to 10 percentage points, red for larger drops.}
\label{tab:summary}
\resizebox{\textwidth}{!}{\small 
\begin{tabular}{lcccc}
\toprule
\textbf{Task} & \textbf{Alg} & \textbf{Resolution} & \textbf{Internal} & \textbf{External} \\
\midrule
\textbf{DR Staging (ICDR)} & HGB & 392 & 0.692 (0.615--0.765) & \cellcolor{highlight} 0.527 (0.478--0.577) \\
\midrule
Scar & LR & 392 & 0.964 (0.926--0.986) & 0.955 (0.923--0.977) \\
Diabetic Retinopathy & LR & 672 & 0.940 (0.916--0.963) & 0.946 (0.932--0.957) \\
Hemorrhage & LR & 672 & 0.804 (0.670--0.891) & \cellcolor{highlight} 0.569 (0.451--0.716) \\
AMD & LR & 952 & 0.976 (0.961--0.987) & \cellcolor{highlight2} 0.909 (0.857--0.948) \\
Nevus & LR & 952 & 0.840 (0.729--0.903) & \cellcolor{highlight2} 0.795 (0.711--0.886) \\
Other & LR & 952 & 0.828 (0.784--0.858) & \cellcolor{highlight2} 0.740 (0.708--0.774) \\
Hypertensive Retinopathy & RF & 392 & 0.864 (0.809--0.909) & \cellcolor{highlight} 0.690 (0.635--0.745) \\
Macular Edema & RF & 392 & 0.952 (0.928--0.975) & 0.952 (0.931--0.965) \\
Vascular Occlusion & RF & 392 & 0.954 (0.913--0.992) & \cellcolor{highlight2} 0.906 (0.835--0.971) \\
Increased Cup/Disc & SVM & 392 & 0.871 (0.855--0.888) & 0.850 (0.836--0.863) \\
Myopic Fundus & SVM & 952 & 0.991 (0.986--0.994) & 0.975 (0.963--0.984) \\
\bottomrule
\end{tabular}%
}
\end{table}

\vspace{2em}
\textbf{Downstream classifier choice has a smaller effect than camera platform.} Classifier performance varied across tasks. Linear and nonlinear classifiers each achieved the best performance on different tasks; no single classifier consistently dominated across tasks. This suggests that, under a frozen retinal foundation-model representation, downstream classifier choice is task-dependent and does not by itself eliminate camera-associated performance differences.

\vspace{1em}
\textbf{\large{4 Discussion \& Conclusions}}

Our results show that camera-associated domain shift persists even when using a frozen retinal foundation model. Performance varied across camera platforms and disease tasks, with some tasks showing little change and others showing substantial reductions in target-camera performance. This heterogeneity suggests that camera shift is not uniform across retinal diseases.

Increasing input resolution and changing the downstream classifier did not consistently eliminate camera-associated performance differences. These findings suggest that higher-resolution inputs or more complex classifiers alone are unlikely to provide a general solution to cross-camera generalization.

Image quality provides an important context for interpreting these results. Inadequate images were more common in the Nikon cohort than in the Canon cohort (18.6\% vs.\ 8.8\%), particularly because of differences in image-field adequacy. Thus, the observed camera effect may reflect both differences in image acquisition and image quality. However, the available data do not permit these factors to be fully disentangled because of the inherent limitations of the dataset.

Overall, our  findings highlight the importance of evaluating retinal AI systems across the imaging platforms encountered in practice. Reporting performance by camera, rather than relying only on aggregate test performance, can reveal failures that may otherwise remain hidden.

This study is limited by the use of two camera platforms within a single dataset and by the confounding between camera platform and image quality. Future work should evaluate additional cameras and datasets and test whether image-quality control or domain-adaptation methods can reduce the observed shift.

Overall, camera-associated domain shift remains an important consideration even with modern retinal foundation models. Robust deployment therefore requires evaluation across both imaging conditions and disease tasks, rather than assuming that a strong representation will generalize across cameras.
 

\bibliographystyle{iclr2025_conference}
\bibliography{retinal}
\newpage 
\appendix

\section{Appendix}
\subsection{Experimental setup}

Table \ref{tab:experimental_factors} summarizes the variables we examined in this work.

\begin{table}[h!]
\centering
\caption{Experimental factors and levels.}
\label{tab:experimental_factors}
\small
\begin{tabular}{llll}
\toprule
\textbf{Factor} & \textbf{Camera} & \textbf{Resolution} & \textbf{Classifier} \\
\midrule
\textbf{Levels} & Canon CR-2, Nikon NF5050 & 224, 392, 672, 952 &
SVM, kNN, HGB, LR, RF \\
\midrule
\textbf{Factor} & \textbf{Tasks / Extractor} & \textbf{Evaluation} & \textbf{Domain} \\
\midrule
\textbf{Levels} & 12 tasks / Frozen RFG & Patient-level & Within / Cross-camera$^{*}$ \\
\bottomrule
\end{tabular}
\end{table}

\begin{figure}[htbp]
    \centering
    \includegraphics[width=.5\textwidth]{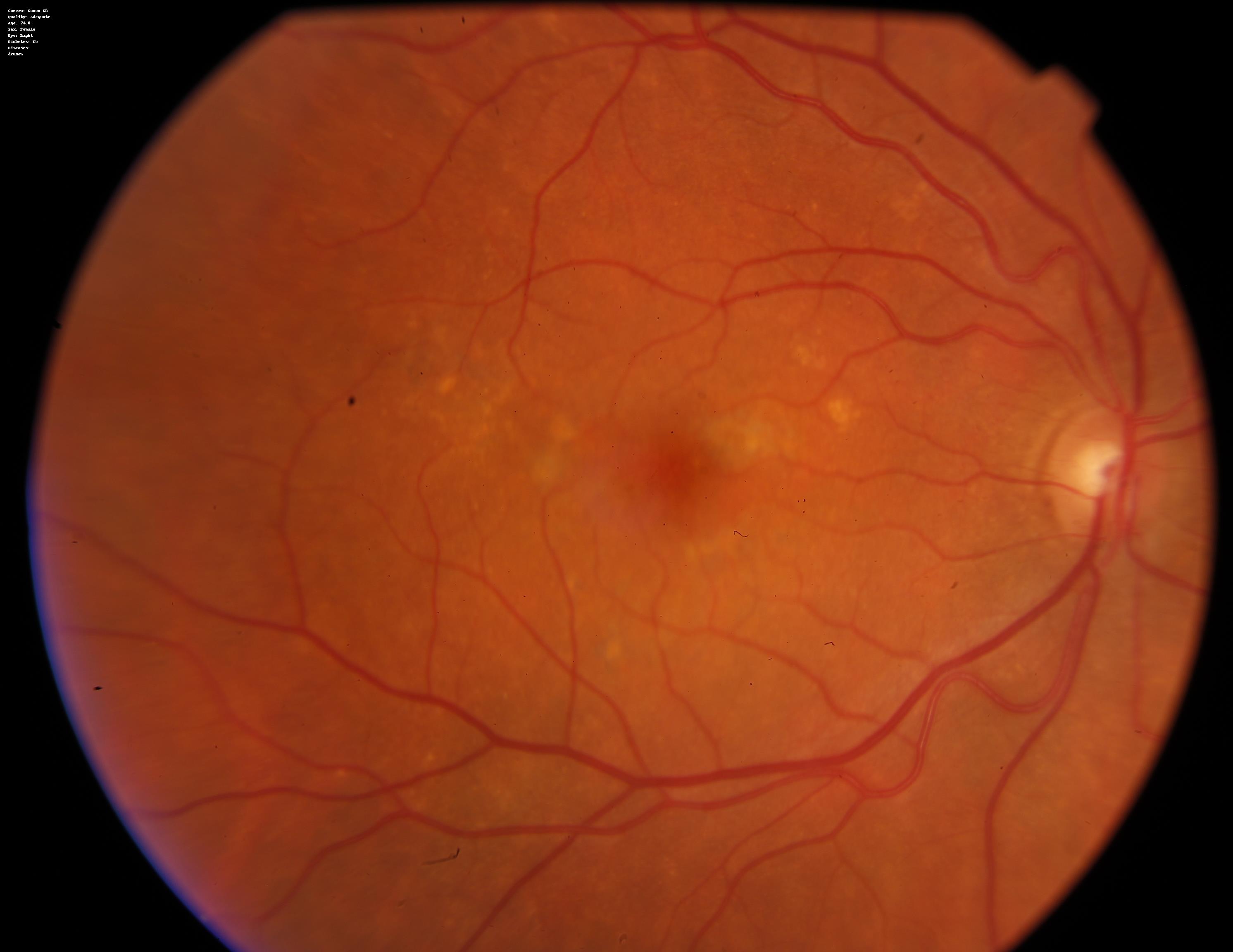}
    \caption{A sample retinal fundus photograph captured using Canon CR-2 for a female patient.}
    \label{fig:cfp}
\end{figure}

\newpage 
\subsection{Results}
{\footnotesize
\begin{longtable}{llccc} 
\caption{Full classifier $\times$ resolution results per task. Internal = held-out Canon test set; External = unseen Nikon cohort. Values show point estimate with 95\% bootstrap confidence interval. Metric is QWK for DR staging (ordinal, 5-class ICDR) and AUROC for all other (binary) tasks.}
\label{tab:full_summary} \\
\toprule 
\textbf{Task} & \textbf{Alg} & \textbf{Resolution} & \textbf{Internal} & \textbf{External} \\
\midrule
\endfirsthead
\multicolumn{5}{c}{\tablename\ \thetable{} -- continued from previous page} \\
\toprule
\textbf{Task} & \textbf{Alg} & \textbf{Resolution} & \textbf{Internal} & \textbf{External} \\
\midrule
\endhead
\midrule
\multicolumn{5}{r}{\textit{Continued on next page}} \\
\endfoot
\bottomrule
\endlastfoot
\multirow{20}{*}{DR Staging (ICDR)} & \multirow{4}{*}{SVM} & 224 & 0.611 (0.508--0.682) & 0.669 (0.619--0.728) \\
 &  & 392 & 0.635 (0.549--0.725) & 0.705 (0.657--0.752) \\
 &  & 672 & 0.662 (0.581--0.739) & 0.716 (0.671--0.760) \\
 &  & 952 & 0.538 (0.452--0.635) & 0.657 (0.610--0.701) \\
\cmidrule(lr){2-5}
 & \multirow{4}{*}{LR} & 224 & 0.579 (0.524--0.629) & 0.601 (0.551--0.652) \\
 &  & 392 & 0.  (0.606--0.699) & 0.579 (0.538--0.632) \\
 &  & 672 & 0.637 (0.580--0.699) & 0.503 (0.469--0.550) \\
 &  & 952 & 0.587 (0.523--0.656) & 0.549 (0.513--0.596) \\
\cmidrule(lr){2-5}
 & \multirow{4}{*}{HGB} & 224 & 0.642 (0.549--0.715) & 0.580 (0.525--0.634) \\
 &  & 392 & 0.692 (0.615--0.765) & 0.527 (0.478--0.577) \\
 &  & 672 & 0.574 (0.499--0.672) & 0.350 (0.284--0.410) \\
 &  & 952 & 0.516 (0.415--0.607) & 0.292 (0.207--0.359) \\
\cmidrule(lr){2-5}
 & \multirow{4}{*}{kNN} & 224 & 0.487 (0.390--0.590) & 0.276 (0.215--0.353) \\
 &  & 392 & 0.469 (0.372--0.584) & 0.229 (0.166--0.291) \\
 &  & 672 & 0.329 (0.216--0.443) & 0.130 (0.076--0.195) \\
 &  & 952 & 0.296 (0.179--0.407) & 0.112 (0.061--0.164) \\
\cmidrule(lr){2-5}
 & \multirow{4}{*}{RF} & 224 & 0.301 (0.192--0.409) & 0.065 (0.026--0.107) \\
 &  & 392 & 0.262 (0.165--0.386) & 0.074 (0.031--0.116) \\
 &  & 672 & 0.203 (0.087--0.310) & 0.018 (-0.001--0.043) \\
 &  & 952 & 0.142 (0.051--0.250) & 0.019 (0.000--0.044) \\
\midrule
\multirow{20}{*}{Diabetic Retinopathy} & \multirow{4}{*}{SVM} & 224 & 0.887 (0.847--0.915) & 0.893 (0.872--0.911) \\
 &  & 392 & 0.933 (0.905--0.954) & 0.930 (0.915--0.946) \\
 &  & 672 & 0.934 (0.906--0.958) & 0.936 (0.918--0.949) \\
 &  & 952 & 0.935 (0.910--0.953) & 0.912 (0.893--0.926) \\
\cmidrule(lr){2-5}
 & \multirow{4}{*}{LR} & 224 & 0.893 (0.859--0.922) & 0.906 (0.886--0.924) \\
 &  & 392 & 0.933 (0.907--0.956) & 0.940 (0.926--0.954) \\
 &  & 672 & 0.940 (0.916--0.963) & 0.946 (0.932--0.957) \\
 &  & 952 & 0.938 (0.912--0.957) & 0.930 (0.914--0.943) \\
\cmidrule(lr){2-5}
 & \multirow{4}{*}{HGB} & 224 & 0.900 (0.869--0.926) & 0.897 (0.878--0.919) \\
 &  & 392 & 0.926 (0.903--0.948) & 0.933 (0.915--0.948) \\
 &  & 672 & 0.919 (0.890--0.941) & 0.924 (0.907--0.941) \\
 &  & 952 & 0.911 (0.890--0.932) & 0.907 (0.893--0.924) \\
\cmidrule(lr){2-5}
 & \multirow{4}{*}{kNN} & 224 & 0.817 (0.778--0.853) & 0.796 (0.763--0.821) \\
 &  & 392 & 0.837 (0.800--0.871) & 0.829 (0.802--0.856) \\
 &  & 672 & 0.808 (0.765--0.853) & 0.794 (0.767--0.822) \\
 &  & 952 & 0.770 (0.723--0.815) & 0.770 (0.745--0.795) \\
\cmidrule(lr){2-5}
 & \multirow{4}{*}{RF} & 224 & 0.894 (0.864--0.918) & 0.891 (0.871--0.911) \\
 &  & 392 & 0.913 (0.883--0.939) & 0.931 (0.917--0.945) \\
 &  & 672 & 0.908 (0.886--0.936) & 0.927 (0.912--0.941) \\
 &  & 952 & 0.889 (0.863--0.914) & 0.893 (0.877--0.908) \\
\midrule
\multirow{20}{*}{AMD} & \multirow{4}{*}{SVM} & 224 & 0.925 (0.887--0.959) & 0.906 (0.870--0.941) \\
 &  & 392 & 0.948 (0.912--0.973) & 0.888 (0.844--0.928) \\
 &  & 672 & 0.971 (0.958--0.984) & 0.889 (0.842--0.932) \\
 &  & 952 & 0.965 (0.936--0.985) & 0.890 (0.829--0.935) \\
\cmidrule(lr){2-5}
 & \multirow{4}{*}{LR} & 224 & 0.941 (0.910--0.968) & 0.907 (0.873--0.937) \\
 &  & 392 & 0.962 (0.939--0.979) & 0.901 (0.865--0.941) \\
 &  & 672 & 0.975 (0.958--0.989) & 0.915 (0.873--0.951) \\
 &  & 952 & 0.976 (0.961--0.987) & 0.909 (0.857--0.948) \\
\cmidrule(lr){2-5}
 & \multirow{4}{*}{HGB} & 224 & 0.958 (0.939--0.973) & 0.821 (0.751--0.870) \\
 &  & 392 & 0.960 (0.938--0.978) & 0.839 (0.779--0.883) \\
 &  & 672 & 0.972 (0.956--0.984) & 0.819 (0.768--0.875) \\
 &  & 952 & 0.967 (0.949--0.982) & 0.831 (0.772--0.891) \\
\cmidrule(lr){2-5}
 & \multirow{4}{*}{kNN} & 224 & 0.794 (0.719--0.847) & 0.671 (0.614--0.725) \\
 &  & 392 & 0.857 (0.786--0.905) & 0.741 (0.682--0.806) \\
 &  & 672 & 0.899 (0.851--0.946) & 0.704 (0.638--0.762) \\
 &  & 952 & 0.849 (0.802--0.899) & 0.737 (0.672--0.795) \\
\cmidrule(lr){2-5}
 & \multirow{4}{*}{RF} & 224 & 0.943 (0.916--0.962) & 0.790 (0.711--0.855) \\
 &  & 392 & 0.954 (0.930--0.974) & 0.829 (0.761--0.877) \\
 &  & 672 & 0.958 (0.933--0.980) & 0.813 (0.744--0.866) \\
 &  & 952 & 0.958 (0.938--0.973) & 0.804 (0.715--0.865) \\
\midrule
\multirow{20}{*}{Hemorrhage} & \multirow{4}{*}{SVM} & 224 & 0.782 (0.619--0.888) & 0.586 (0.454--0.710) \\
 &  & 392 & 0.709 (0.559--0.797) & 0.568 (0.453--0.700) \\
 &  & 672 & 0.779 (0.657--0.867) & 0.555 (0.432--0.685) \\
 &  & 952 & 0.705 (0.560--0.808) & 0.572 (0.470--0.697) \\
\cmidrule(lr){2-5}
 & \multirow{4}{*}{LR} & 224 & 0.794 (0.619--0.906) & 0.601 (0.478--0.733) \\
 &  & 392 & 0.754 (0.599--0.840) & 0.580 (0.456--0.707) \\
 &  & 672 & 0.804 (0.670--0.891) & 0.569 (0.451--0.716) \\
 &  & 952 & 0.728 (0.565--0.841) & 0.594 (0.491--0.723) \\
\cmidrule(lr){2-5}
 & \multirow{4}{*}{HGB} & 224 & 0.705 (0.506--0.824) & 0.682 (0.552--0.784) \\
 &  & 392 & 0.757 (0.537--0.887) & 0.610 (0.511--0.738) \\
 &  & 672 & 0.738 (0.584--0.884) & 0.676 (0.579--0.794) \\
 &  & 952 & 0.701 (0.485--0.864) & 0.674 (0.540--0.775) \\
\cmidrule(lr){2-5}
 & \multirow{4}{*}{kNN} & 224 & 0.622 (0.519--0.733) & 0.551 (0.480--0.624) \\
 &  & 392 & 0.596 (0.478--0.694) & 0.583 (0.499--0.665) \\
 &  & 672 & 0.500 (0.463--0.550) & 0.534 (0.478--0.612) \\
 &  & 952 & 0.533 (0.469--0.620) & 0.569 (0.502--0.648) \\
\cmidrule(lr){2-5}
 & \multirow{4}{*}{RF} & 224 & 0.785 (0.621--0.907) & 0.749 (0.670--0.806) \\
 &  & 392 & 0.785 (0.593--0.907) & 0.727 (0.630--0.823) \\
 &  & 672 & 0.725 (0.516--0.884) & 0.696 (0.558--0.824) \\
 &  & 952 & 0.707 (0.560--0.848) & 0.746 (0.627--0.846) \\
\midrule
\multirow{20}{*}{Hypertensive Retinopathy} & \multirow{4}{*}{SVM} & 224 & 0.703 (0.597--0.805) & 0.530 (0.470--0.579) \\
 &  & 392 & 0.793 (0.719--0.877) & 0.523 (0.462--0.567) \\
 &  & 672 & 0.776 (0.682--0.857) & 0.567 (0.507--0.618) \\
 &  & 952 & 0.776 (0.688--0.864) & 0.532 (0.478--0.586) \\
\cmidrule(lr){2-5}
 & \multirow{4}{*}{LR} & 224 & 0.734 (0.645--0.836) & 0.568 (0.511--0.621) \\
 &  & 392 & 0.840 (0.774--0.914) & 0.564 (0.502--0.609) \\
 &  & 672 & 0.822 (0.739--0.901) & 0.599 (0.546--0.648) \\
 &  & 952 & 0.803 (0.721--0.879) & 0.552 (0.494--0.601) \\
\cmidrule(lr){2-5}
 & \multirow{4}{*}{HGB} & 224 & 0.818 (0.747--0.889) & 0.664 (0.625--0.708) \\
 &  & 392 & 0.826 (0.741--0.891) & 0.663 (0.601--0.710) \\
 &  & 672 & 0.799 (0.723--0.856) & 0.673 (0.617--0.715) \\
 &  & 952 & 0.761 (0.696--0.832) & 0.634 (0.586--0.698) \\
\cmidrule(lr){2-5}
 & \multirow{4}{*}{kNN} & 224 & 0.646 (0.575--0.732) & 0.547 (0.503--0.607) \\
 &  & 392 & 0.673 (0.607--0.762) & 0.568 (0.530--0.607) \\
 &  & 672 & 0.616 (0.554--0.698) & 0.517 (0.466--0.567) \\
 &  & 952 & 0.617 (0.545--0.710) & 0.502 (0.462--0.551) \\
\cmidrule(lr){2-5}
 & \multirow{4}{*}{RF} & 224 & 0.797 (0.694--0.870) & 0.690 (0.629--0.753) \\
 &  & 392 & 0.864 (0.809--0.909) & 0.690 (0.635--0.745) \\
 &  & 672 & 0.820 (0.731--0.875) & 0.622 (0.579--0.672) \\
 &  & 952 & 0.794 (0.725--0.848) & 0.628 (0.587--0.679) \\
\midrule
\multirow{20}{*}{Increased Cup/Disc} & \multirow{4}{*}{SVM} & 224 & 0.853 (0.836--0.869) & 0.829 (0.817--0.841) \\
 &  & 392 & 0.871 (0.855--0.888) & 0.850 (0.836--0.863) \\
 &  & 672 & 0.865 (0.848--0.879) & 0.844 (0.827--0.855) \\
 &  & 952 & 0.859 (0.838--0.872) & 0.843 (0.829--0.855) \\
\cmidrule(lr){2-5}
 & \multirow{4}{*}{LR} & 224 & 0.854 (0.838--0.872) & 0.839 (0.828--0.850) \\
 &  & 392 & 0.870 (0.854--0.887) & 0.856 (0.843--0.869) \\
 &  & 672 & 0.868 (0.851--0.882) & 0.852 (0.837--0.864) \\
 &  & 952 & 0.862 (0.841--0.874) & 0.846 (0.831--0.857) \\
\cmidrule(lr){2-5}
 & \multirow{4}{*}{HGB} & 224 & 0.804 (0.781--0.823) & 0.749 (0.730--0.767) \\
 &  & 392 & 0.813 (0.790--0.830) & 0.793 (0.777--0.807) \\
 &  & 672 & 0.818 (0.800--0.835) & 0.783 (0.766--0.799) \\
 &  & 952 & 0.819 (0.799--0.836) & 0.781 (0.762--0.795) \\
\cmidrule(lr){2-5}
 & \multirow{4}{*}{kNN} & 224 & 0.737 (0.710--0.760) & 0.683 (0.662--0.703) \\
 &  & 392 & 0.729 (0.700--0.751) & 0.699 (0.676--0.714) \\
 &  & 672 & 0.735 (0.715--0.758) & 0.678 (0.658--0.695) \\
 &  & 952 & 0.719 (0.694--0.739) & 0.664 (0.642--0.684) \\
\cmidrule(lr){2-5}
 & \multirow{4}{*}{RF} & 224 & 0.782 (0.759--0.801) & 0.746 (0.730--0.766) \\
 &  & 392 & 0.790 (0.773--0.808) & 0.784 (0.767--0.801) \\
 &  & 672 & 0.793 (0.771--0.808) & 0.782 (0.765--0.798) \\
 &  & 952 & 0.792 (0.771--0.808) & 0.783 (0.767--0.798) \\
\midrule
\multirow{20}{*}{Macular Edema} & \multirow{4}{*}{SVM} & 224 & 0.881 (0.820--0.949) & 0.961 (0.941--0.978) \\
 &  & 392 & 0.897 (0.815--0.969) & 0.984 (0.975--0.991) \\
 &  & 672 & 0.916 (0.856--0.973) & 0.983 (0.976--0.989) \\
 &  & 952 & 0.917 (0.867--0.970) & 0.979 (0.965--0.989) \\
\cmidrule(lr){2-5}
 & \multirow{4}{*}{LR} & 224 & 0.906 (0.853--0.962) & 0.968 (0.948--0.985) \\
 &  & 392 & 0.913 (0.839--0.974) & 0.987 (0.980--0.994) \\
 &  & 672 & 0.927 (0.875--0.978) & 0.990 (0.986--0.993) \\
 &  & 952 & 0.928 (0.883--0.977) & 0.982 (0.970--0.992) \\
\cmidrule(lr){2-5}
 & \multirow{4}{*}{HGB} & 224 & 0.946 (0.919--0.969) & 0.946 (0.929--0.959) \\
 &  & 392 & 0.934 (0.895--0.971) & 0.965 (0.951--0.978) \\
 &  & 672 & 0.945 (0.909--0.976) & 0.962 (0.949--0.974) \\
 &  & 952 & 0.939 (0.911--0.966) & 0.963 (0.953--0.974) \\
\cmidrule(lr){2-5}
 & \multirow{4}{*}{kNN} & 224 & 0.809 (0.724--0.878) & 0.851 (0.816--0.888) \\
 &  & 392 & 0.844 (0.781--0.893) & 0.901 (0.871--0.936) \\
 &  & 672 & 0.765 (0.678--0.847) & 0.842 (0.808--0.892) \\
 &  & 952 & 0.750 (0.671--0.814) & 0.786 (0.752--0.840) \\
\cmidrule(lr){2-5}
 & \multirow{4}{*}{RF} & 224 & 0.942 (0.917--0.964) & 0.936 (0.916--0.959) \\
 &  & 392 & 0.952 (0.928--0.975) & 0.952 (0.931--0.965) \\
 &  & 672 & 0.948 (0.920--0.970) & 0.954 (0.936--0.969) \\
 &  & 952 & 0.926 (0.889--0.955) & 0.954 (0.935--0.969) \\
\midrule
\multirow{20}{*}{Myopic Fundus} & \multirow{4}{*}{SVM} & 224 & 0.960 (0.923--0.986) & 0.975 (0.965--0.983) \\
 &  & 392 & 0.974 (0.953--0.991) & 0.978 (0.966--0.989) \\
 &  & 672 & 0.990 (0.984--0.994) & 0.972 (0.959--0.986) \\
 &  & 952 & 0.991 (0.986--0.994) & 0.975 (0.963--0.984) \\
\cmidrule(lr){2-5}
 & \multirow{4}{*}{LR} & 224 & 0.970 (0.944--0.990) & 0.985 (0.979--0.991) \\
 &  & 392 & 0.981 (0.965--0.992) & 0.983 (0.972--0.991) \\
 &  & 672 & 0.989 (0.983--0.994) & 0.978 (0.958--0.992) \\
 &  & 952 & 0.989 (0.984--0.993) & 0.977 (0.968--0.989) \\
\cmidrule(lr){2-5}
 & \multirow{4}{*}{HGB} & 224 & 0.981 (0.970--0.991) & 0.965 (0.942--0.986) \\
 &  & 392 & 0.986 (0.978--0.993) & 0.990 (0.984--0.994) \\
 &  & 672 & 0.982 (0.973--0.989) & 0.975 (0.955--0.989) \\
 &  & 952 & 0.983 (0.971--0.990) & 0.972 (0.952--0.989) \\
\cmidrule(lr){2-5}
 & \multirow{4}{*}{kNN} & 224 & 0.905 (0.838--0.960) & 0.878 (0.835--0.925) \\
 &  & 392 & 0.890 (0.826--0.955) & 0.864 (0.817--0.927) \\
 &  & 672 & 0.892 (0.827--0.951) & 0.882 (0.841--0.924) \\
 &  & 952 & 0.892 (0.827--0.950) & 0.835 (0.782--0.888) \\
\cmidrule(lr){2-5}
 & \multirow{4}{*}{RF} & 224 & 0.980 (0.971--0.990) & 0.975 (0.955--0.991) \\
 &  & 392 & 0.983 (0.975--0.991) & 0.989 (0.982--0.994) \\
 &  & 672 & 0.983 (0.975--0.991) & 0.986 (0.979--0.993) \\
 &  & 952 & 0.981 (0.972--0.989) & 0.981 (0.969--0.990) \\
\midrule
\multirow{20}{*}{Nevus} & \multirow{4}{*}{SVM} & 224 & 0.721 (0.614--0.854) & 0.724 (0.632--0.815) \\
 &  & 392 & 0.685 (0.578--0.803) & 0.751 (0.654--0.839) \\
 &  & 672 & 0.822 (0.749--0.878) & 0.771 (0.682--0.861) \\
 &  & 952 & 0.830 (0.712--0.893) & 0.795 (0.710--0.885) \\
\cmidrule(lr){2-5}
 & \multirow{4}{*}{LR} & 224 & 0.724 (0.610--0.847) & 0.738 (0.659--0.831) \\
 &  & 392 & 0.708 (0.601--0.827) & 0.754 (0.662--0.842) \\
 &  & 672 & 0.832 (0.772--0.894) & 0.775 (0.692--0.858) \\
 &  & 952 & 0.840 (0.729--0.903) & 0.795 (0.711--0.886) \\
\cmidrule(lr){2-5}
 & \multirow{4}{*}{HGB} & 224 & 0.521 (0.380--0.643) & 0.617 (0.517--0.703) \\
 &  & 392 & 0.582 (0.435--0.694) & 0.683 (0.582--0.781) \\
 &  & 672 & 0.607 (0.503--0.731) & 0.666 (0.570--0.746) \\
 &  & 952 & 0.526 (0.364--0.646) & 0.591 (0.490--0.687) \\
\cmidrule(lr){2-5}
 & \multirow{4}{*}{kNN} & 224 & 0.553 (0.480--0.658) & 0.520 (0.461--0.593) \\
 &  & 392 & 0.473 (0.424--0.549) & 0.538 (0.469--0.605) \\
 &  & 672 & 0.514 (0.456--0.602) & 0.546 (0.477--0.607) \\
 &  & 952 & 0.490 (0.439--0.562) & 0.593 (0.508--0.671) \\
\cmidrule(lr){2-5}
 & \multirow{4}{*}{RF} & 224 & 0.548 (0.396--0.659) & 0.610 (0.493--0.720) \\
 &  & 392 & 0.493 (0.387--0.592) & 0.644 (0.544--0.740) \\
 &  & 672 & 0.563 (0.423--0.670) & 0.634 (0.529--0.725) \\
 &  & 952 & 0.539 (0.401--0.647) & 0.548 (0.451--0.650) \\
\midrule
\multirow{20}{*}{Other} & \multirow{4}{*}{SVM} & 224 & 0.808 (0.770--0.842) & 0.695 (0.659--0.732) \\
 &  & 392 & 0.789 (0.725--0.828) & 0.720 (0.692--0.750) \\
 &  & 672 & 0.796 (0.748--0.832) & 0.695 (0.663--0.731) \\
 &  & 952 & 0.808 (0.757--0.840) & 0.712 (0.680--0.747) \\
\cmidrule(lr){2-5}
 & \multirow{4}{*}{LR} & 224 & 0.822 (0.785--0.852) & 0.729 (0.694--0.762) \\
 &  & 392 & 0.804 (0.750--0.845) & 0.749 (0.721--0.781) \\
 &  & 672 & 0.812 (0.768--0.847) & 0.727 (0.698--0.758) \\
 &  & 952 & 0.828 (0.784--0.858) & 0.740 (0.708--0.774) \\
\cmidrule(lr){2-5}
 & \multirow{4}{*}{HGB} & 224 & 0.796 (0.753--0.827) & 0.697 (0.669--0.724) \\
 &  & 392 & 0.785 (0.751--0.819) & 0.722 (0.700--0.745) \\
 &  & 672 & 0.797 (0.766--0.832) & 0.736 (0.706--0.768) \\
 &  & 952 & 0.803 (0.767--0.832) & 0.732 (0.706--0.765) \\
\cmidrule(lr){2-5}
 & \multirow{4}{*}{kNN} & 224 & 0.673 (0.620--0.720) & 0.595 (0.564--0.630) \\
 &  & 392 & 0.671 (0.623--0.716) & 0.625 (0.599--0.657) \\
 &  & 672 & 0.660 (0.606--0.703) & 0.614 (0.581--0.641) \\
 &  & 952 & 0.615 (0.563--0.670) & 0.606 (0.577--0.636) \\
\cmidrule(lr){2-5}
 & \multirow{4}{*}{RF} & 224 & 0.790 (0.750--0.824) & 0.732 (0.700--0.761) \\
 &  & 392 & 0.789 (0.759--0.822) & 0.735 (0.709--0.761) \\
 &  & 672 & 0.782 (0.745--0.816) & 0.746 (0.722--0.778) \\
 &  & 952 & 0.798 (0.761--0.826) & 0.749 (0.718--0.780) \\
\midrule
\multirow{20}{*}{Scar} & \multirow{4}{*}{SVM} & 224 & 0.927 (0.849--0.979) & 0.961 (0.943--0.976) \\
 &  & 392 & 0.948 (0.895--0.977) & 0.947 (0.916--0.973) \\
 &  & 672 & 0.955 (0.920--0.977) & 0.943 (0.908--0.969) \\
 &  & 952 & 0.942 (0.892--0.973) & 0.953 (0.924--0.976) \\
\cmidrule(lr){2-5}
 & \multirow{4}{*}{LR} & 224 & 0.945 (0.887--0.989) & 0.958 (0.937--0.979) \\
 &  & 392 & 0.964 (0.926--0.986) & 0.955 (0.923--0.977) \\
 &  & 672 & 0.959 (0.915--0.982) & 0.957 (0.925--0.977) \\
 &  & 952 & 0.952 (0.910--0.982) & 0.957 (0.934--0.978) \\
\cmidrule(lr){2-5}
 & \multirow{4}{*}{HGB} & 224 & 0.925 (0.861--0.974) & 0.938 (0.911--0.961) \\
 &  & 392 & 0.927 (0.838--0.971) & 0.910 (0.867--0.939) \\
 &  & 672 & 0.928 (0.876--0.963) & 0.930 (0.892--0.956) \\
 &  & 952 & 0.925 (0.837--0.974) & 0.930 (0.892--0.960) \\
\cmidrule(lr){2-5}
 & \multirow{4}{*}{kNN} & 224 & 0.769 (0.675--0.857) & 0.815 (0.767--0.863) \\
 &  & 392 & 0.781 (0.683--0.864) & 0.816 (0.759--0.854) \\
 &  & 672 & 0.754 (0.677--0.841) & 0.781 (0.731--0.827) \\
 &  & 952 & 0.737 (0.661--0.820) & 0.768 (0.718--0.822) \\
\cmidrule(lr){2-5}
 & \multirow{4}{*}{RF} & 224 & 0.921 (0.854--0.963) & 0.920 (0.892--0.948) \\
 &  & 392 & 0.923 (0.833--0.968) & 0.897 (0.844--0.932) \\
 &  & 672 & 0.919 (0.827--0.969) & 0.917 (0.869--0.945) \\
 &  & 952 & 0.920 (0.843--0.967) & 0.928 (0.888--0.957) \\
\midrule
\multirow{20}{*}{Vascular Occlusion} & \multirow{4}{*}{SVM} & 224 & 0.913 (0.834--0.968) & 0.935 (0.889--0.978) \\
 &  & 392 & 0.929 (0.865--0.968) & 0.919 (0.831--0.985) \\
 &  & 672 & 0.913 (0.824--0.975) & 0.932 (0.870--0.982) \\
 &  & 952 & 0.872 (0.773--0.946) & 0.921 (0.861--0.973) \\
\cmidrule(lr){2-5}
 & \multirow{4}{*}{LR} & 224 & 0.923 (0.853--0.967) & 0.943 (0.906--0.980) \\
 &  & 392 & 0.950 (0.904--0.978) & 0.930 (0.857--0.988) \\
 &  & 672 & 0.917 (0.833--0.970) & 0.943 (0.886--0.985) \\
 &  & 952 & 0.894 (0.809--0.955) & 0.934 (0.877--0.978) \\
\cmidrule(lr){2-5}
 & \multirow{4}{*}{HGB} & 224 & 0.938 (0.892--0.973) & 0.814 (0.716--0.937) \\
 &  & 392 & 0.949 (0.914--0.990) & 0.878 (0.800--0.960) \\
 &  & 672 & 0.902 (0.824--0.970) & 0.784 (0.665--0.897) \\
 &  & 952 & 0.794 (0.653--0.907) & 0.791 (0.693--0.902) \\
\cmidrule(lr){2-5}
 & \multirow{4}{*}{kNN} & 224 & 0.833 (0.714--0.934) & 0.778 (0.679--0.906) \\
 &  & 392 & 0.831 (0.722--0.932) & 0.769 (0.647--0.891) \\
 &  & 672 & 0.808 (0.683--0.906) & 0.732 (0.645--0.820) \\
 &  & 952 & 0.740 (0.607--0.849) & 0.769 (0.665--0.878) \\
\cmidrule(lr){2-5}
 & \multirow{4}{*}{RF} & 224 & 0.941 (0.863--0.990) & 0.902 (0.826--0.966) \\
 &  & 392 & 0.954 (0.913--0.992) & 0.906 (0.835--0.971) \\
 &  & 672 & 0.920 (0.840--0.987) & 0.934 (0.887--0.973) \\
 &  & 952 & 0.930 (0.862--0.977) & 0.868 (0.767--0.941) \\
\end{longtable}
}

\begin{figure}[htbp]
    \centering
    \includegraphics[width=1.\textwidth]{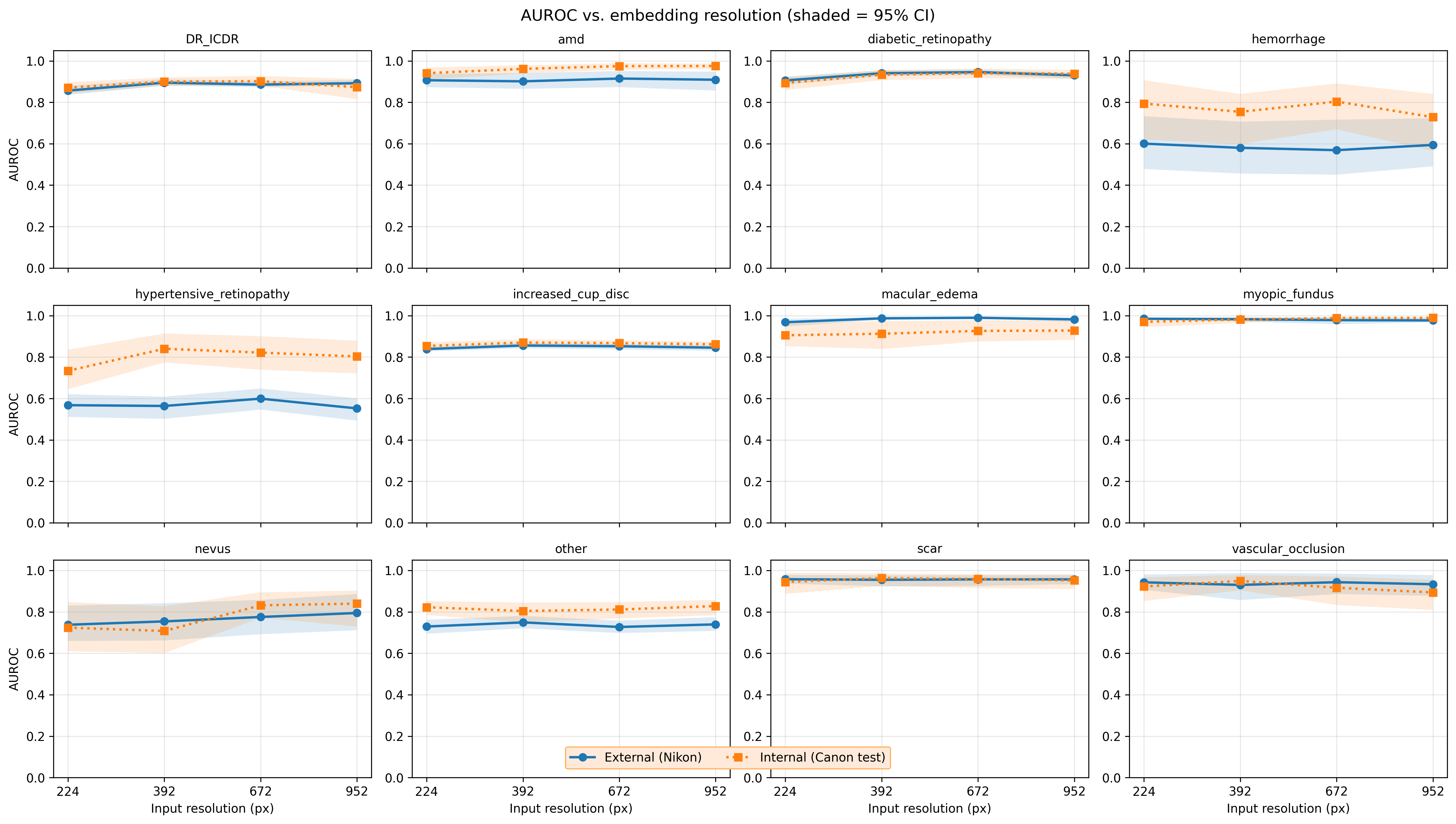}
    \caption{Effect of input resolution on cross-camera generalization for 12 BRSET predictive tasks. Area under the receiver operating characteristic curve (AUROC) for a \textbf{Logistic Regression} trained on frozen RetFound-Green (RFG) embeddings, evaluated at four input resolutions (224, 392, 672, 952 px). Internal (dotted orange) denotes the held-out Canon test split (patient-disjoint from training; n=2,191); External (solid blue) denotes the Nikon subset (n=5,639), representing a distinct camera never seen during training. Shaded regions show 95\% confidence intervals from 100 bootstrap resamples. Most binary pathology labels (e.g., diabetic retinopathy, macular edema, myopic fundus) show overlapping internal/external intervals across resolutions, indicating that RFG embeddings support camera-invariant discrimination for these tasks. In contrast, hemorrhage and hypertensive retinopathy show consistent, non-overlapping degradation under external evaluation ($\Delta$ AUROC=0.15–0.28), suggesting these labels are more sensitive to camera-specific color/optical characteristics than to the classifier itself}
    \label{fig:ref}
\end{figure}

\begin{figure}[htbp]
    \centering
    \includegraphics[width=1\textwidth]{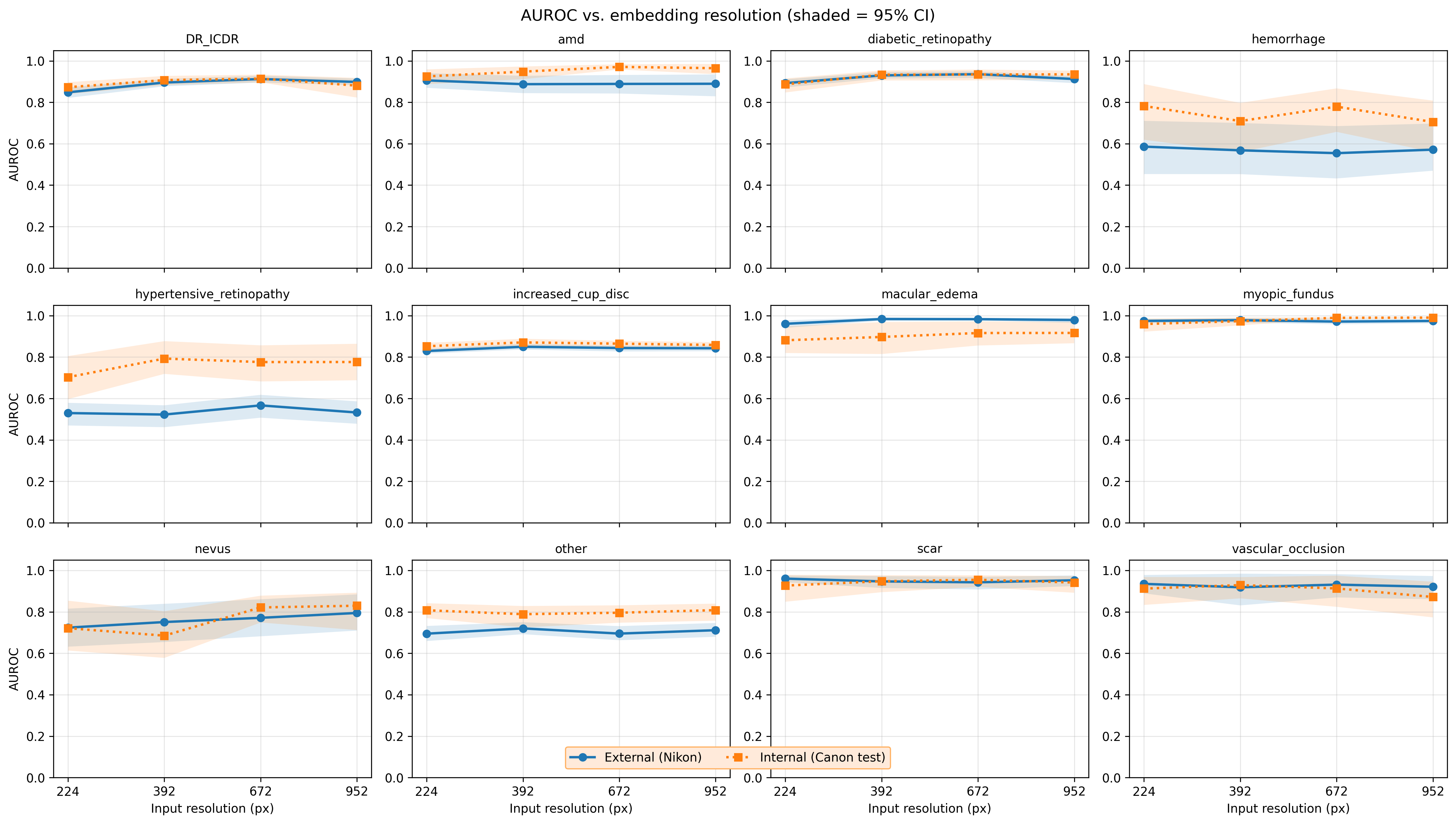}
    \caption{Effects if using a \textbf{Support Vector Machine}. Same setup as Figure \ref{fig:ref} (frozen RFG embeddings, 4 resolutions, internal Canon test vs. external Nikon cohort, 95\% bootstrap CIs). }
    \label{fig:svm}
\end{figure}

\begin{figure}[htbp]
    \centering
    \includegraphics[width=1\textwidth]{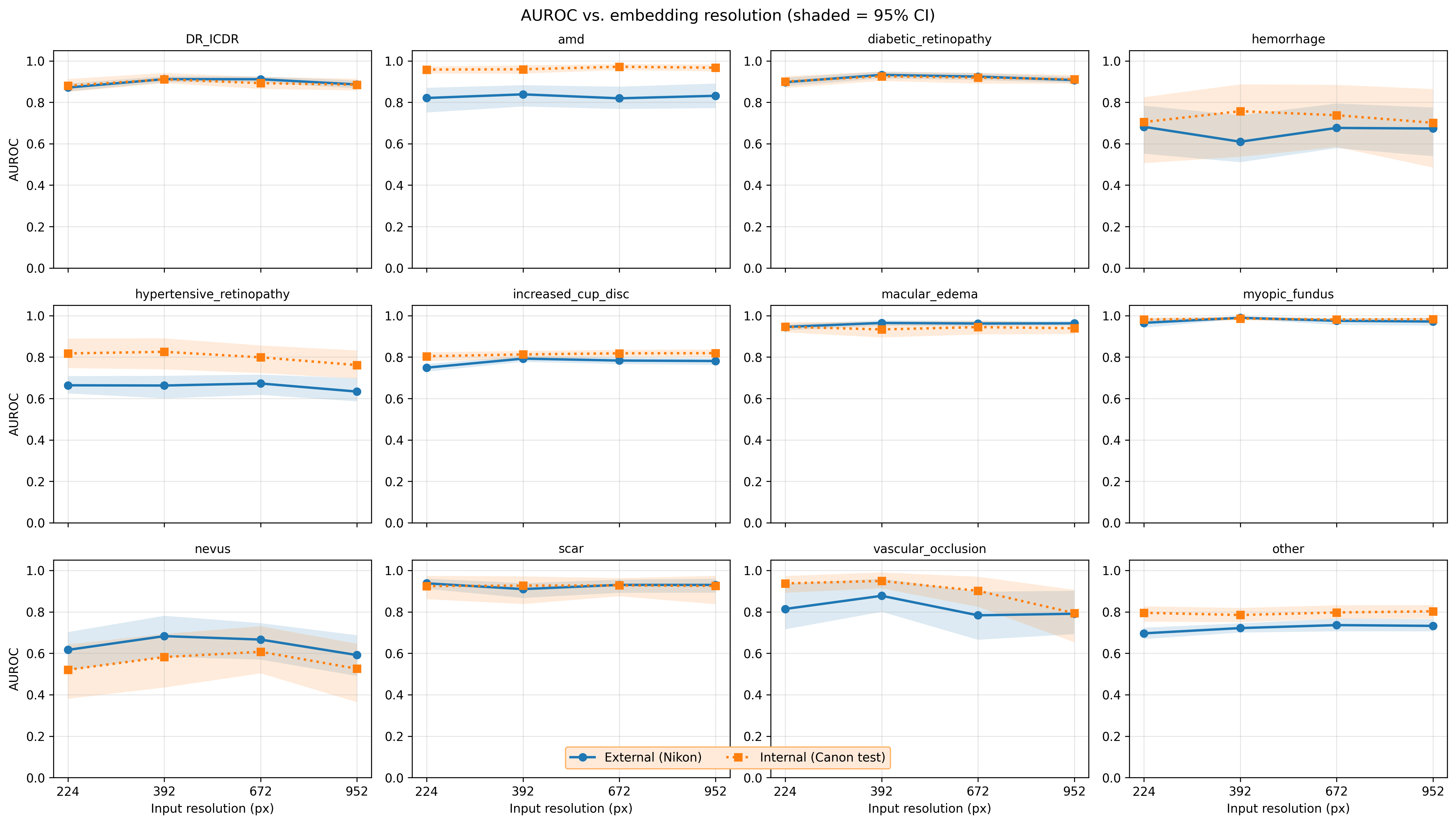}
    \caption{Effects if using \textbf{HGB}. Same setup as Figure \ref{fig:ref} (frozen RFG embeddings, 4 resolutions, internal Canon test vs. external Nikon cohort, 95\% bootstrap CIs). }
    \label{fig:hgb}
\end{figure}

\begin{figure}[htbp]
    \centering
    \includegraphics[width=1\textwidth]{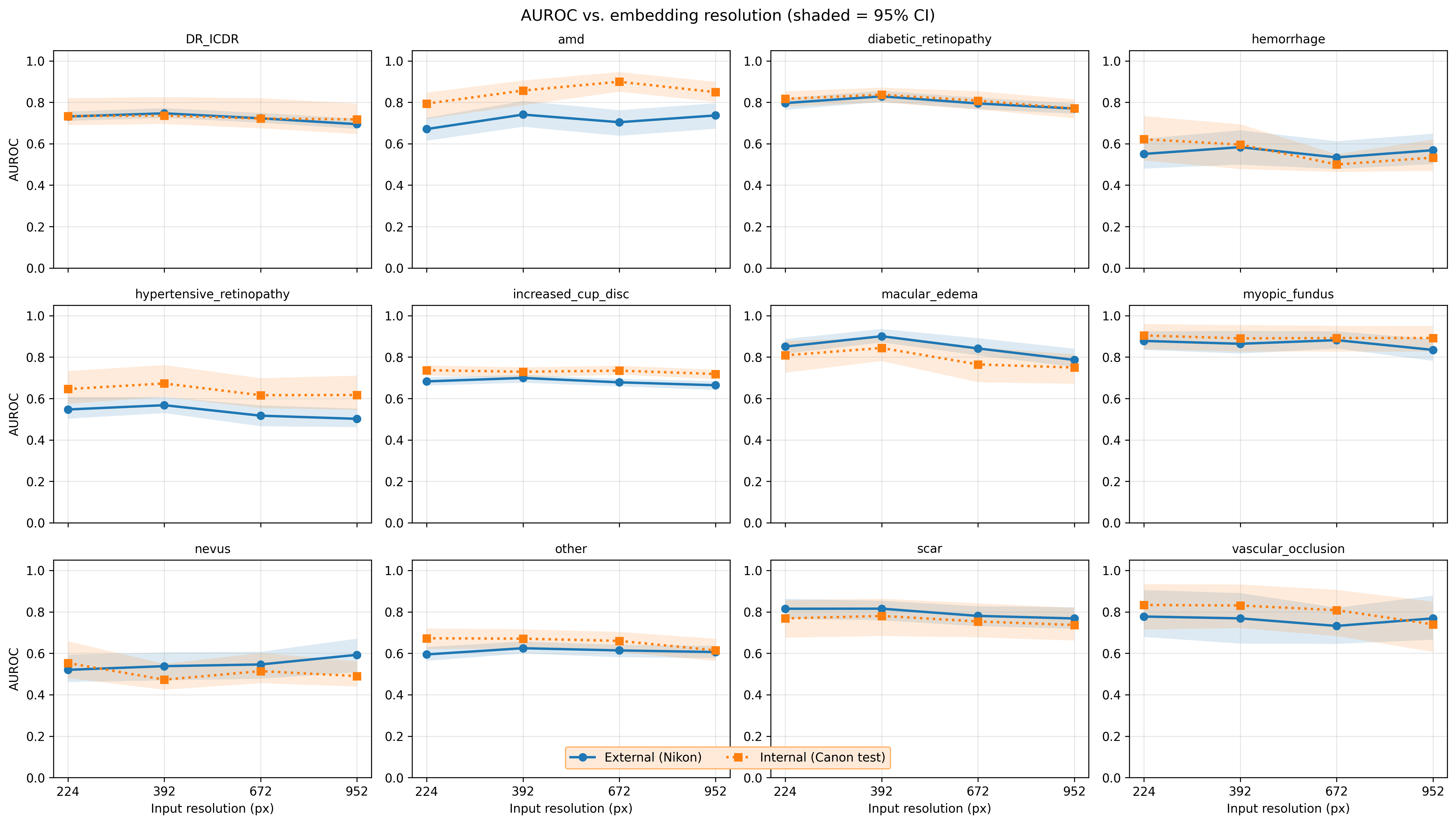}
    \caption{Effects if using \textbf{kNN}. Same setup as Figure \ref{fig:ref} (frozen RFG embeddings, 4 resolutions, internal Canon test vs. external Nikon cohort, 95\% bootstrap CIs). }
    \label{fig:knn}
\end{figure}

\end{document}